\documentclass[10pt,twocolumn,letterpaper]{article}

\usepackage{cvpr}
\usepackage{times}
\usepackage{epsfig}
\usepackage{graphicx}
\usepackage{amsmath}
\usepackage{amssymb}
\usepackage{multirow}
\usepackage{caption}

\DeclareRobustCommand\onedotojw{\futurelet\@let@token\onedotojwaux}
\def\onedotojwaux{\ifx\@let@token.\else.\null\fi\xspace}

\providecommand{\etal}{\emph{et al}\onedotojw}
\newcommand{\fig}[1]{Fig.~\ref{fig:#1}}

\usepackage{authblk}
\usepackage[pagebackref=true,breaklinks=true,letterpaper=true,colorlinks,bookmarks=false]{hyperref}

\cvprfinalcopy 

\ifcvprfinal\fi
\begin{document}

\title{3D Guided Fine-Grained Face Manipulation \vspace{-10pt}}

\setlength{\affilsep}{0.3em}

\author[1,2]{Zhenglin Geng}
\author[2]{Chen Cao}
\author[2]{Sergey Tulyakov}
\affil[1]{Stanford University}
\affil[2]{Snap Inc.}

\makeatletter
\renewcommand\AB@emaillist{\small{\texttt{zhenglin@stanford.edu,\{chen.cao,stulyakov\}@snapchat.com}}}
\makeatother

\twocolumn[{%
\renewcommand\twocolumn[1][]{#1}%
\maketitle
\vspace*{-12mm}
\begin{center}
    \centering
    \includegraphics[width=1.0\linewidth]{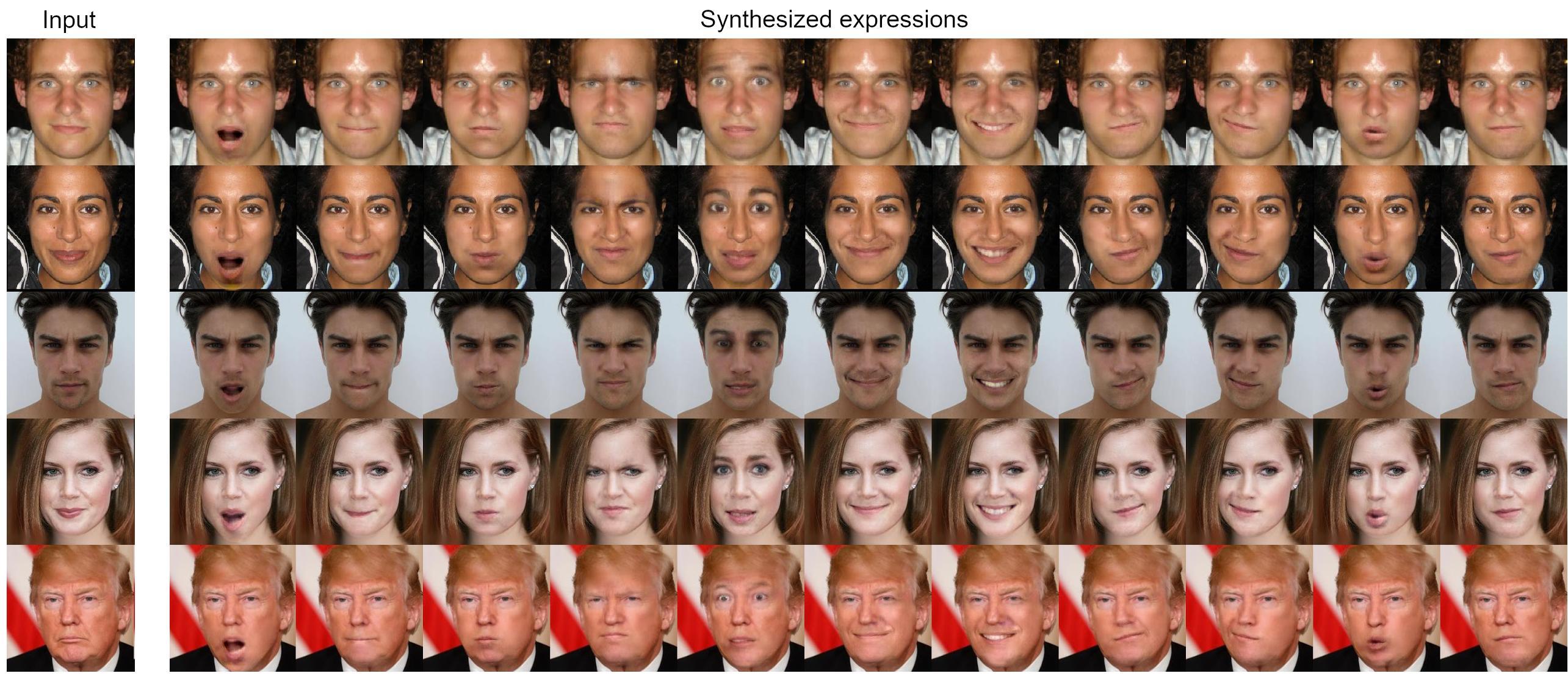}
    \captionof{figure}{\textbf{Qualitative samples.} Given an image, our method can generate multiple realistic expressions of the same subject. }
    \label{fig:teaser}
\end{center}%
}]


\begin{abstract}
We present a method for fine-grained face manipulation. Given a face image with an arbitrary expression, our method can synthesize another arbitrary expression by the same person. 
This is achieved by first fitting a 3D
face model and then disentangling the face into a texture and a shape. We then learn different networks in these two spaces. In the texture space, we use a conditional generative network to change the appearance, and carefully design input formats and loss functions to achieve the best results. In the shape space, we use a fully connected network to predict the accurate shapes and use the available depth data for supervision. Both networks are conditioned on expression coefficients rather than discrete labels, allowing us to generate an unlimited amount of expressions. We show the superiority of this disentangling approach through both quantitative and qualitative studies.
In a user study, our method is preferred in 85\% of cases when compared to the most recent work. When compared to the ground truth, annotators cannot reliably distinguish between our synthesized images and real images, preferring our method in 53\% of the cases.
\end{abstract}

\vspace{-1.2cm}
\section{Introduction}

Face manipulation, a problem involving changing the facial expressions in images enables many creative applications. Until very recently, this problem was mainly addressed from a graphical perspective in which a 3D Morphable Model (3DMM) was first fitted to the image and then re-rendered with a different facial expression. Such techniques jointly model both the shape and the appearance and are typically trained using spatially aligned 3D scans of people \cite{3DMM1999}. A desired facial expression can then be generated by combining graphical primitives called blendshapes~\cite{Li2010}. The blendshapes often  correspond to the Facial Action Coding System (FACS)~\cite{friesen1978facial} which defines a set of anatomically related muscle activations. Unfortunately, due to the Gaussian assumption, 3DMMs often produce blurry shapes and appearances, preventing realistic face rendering.

Deep generative techniques offer a different way of solving the face manipulation problem. In contrast to 3DMMs, they learn an internal representation that jointly models the shape and the appearance of the faces. Manipulation is then performed by conditioning the decoder on expression labels ~\cite{StarGAN2018,lample2017fader} or latent vectors \cite{liu2017unsupervised}. This solution is sub-optimal in several respects. First, neural networks have been recently shown to have difficulties in generating simple geometric transformations~\cite{liu2018intriguing}, whereas face manipulation involves many such transformations, such as mouths opening, eyes closing and other transformations. Second, their models require many examples of such transformations along with their intensities at the  training time, which becomes even more problematic for less common expressions such as sad-smile or negative-surprise. Third, each model supports only a small set of manipulation operations, not allowing fine-grained 3D manipulation.

In this paper we present a novel method that combines 3DMMs and deep generative techniques in a single framework for fine-grained face manipulation. Randomly selected qualitative samples produced by our method are given in \fig{teaser}. Given a face image, we first fit a 3D face model on the image to obtain the texture and the shape. The shape is further represented as identity and expression coefficients using a bilinear model~\cite{cao2014facewarehouse}. This way we disentangle the shape and the texture spaces and use separate branches in our pipeline to apply transformations in these spaces.

The texture branch consists of a convolutional neural network and assumes the texture and the desired expression as inputs, producing a new texture which corresponds to the desired expression. Due to the difficulties of the convolutional networks in generating geometric transformations, we propose conditioning the texture branch on the UV maps that describes target geometry information instead of directly concatenating the labels as in~\cite{StarGAN2018} or coefficients as in~\cite{GANImation2018}. To better preserve texture-expression consistency and the identities in the generated images, we design corresponding loss functions for improved results.  

The shape branch is implemented using a fully connected neural network taking the identity and the expression coefficients as inputs and outputting \emph{shape deformation} necessary to accurately match the desired expression. Notably, a common problem in fitting a morphable model to the face is its inability to fully capture the face shape given only a 2D RGB input image sparsely labeled with 2D landmarks. This is often called face shape hallucination~\cite{tulyakov2018consistent}. At training time, to improve 3D reconstruction, we additionally supervise the shape branch using the available depth data in the FaceWarehouse dataset~\cite{cao2014facewarehouse}.

The proposed approach has a number of benefits. First, we disentangle the texture and shape shapes to make it easier to learn for each branch. In the texture space, faces tend to be more similar despite significant variance in the image space caused by different poses and expressions. Therefore, the texture branch only focuses on the appearance details such as wrinkles, shadows and shading. Similarly, the shape branch focuses on the geometric details only. Second, since we represent expressions as a combination of Face Action Unit coefficients~\cite{friesen1978facial}, rather than discrete labels, our approach can generate infinite number of target expressions. Third, we further distinguish identity and expression coefficients, to better preserve subject-specific features by only changing the expression components in the shape space.

\begin{figure*}[t]
    \centering
    \includegraphics[width=0.95\linewidth]{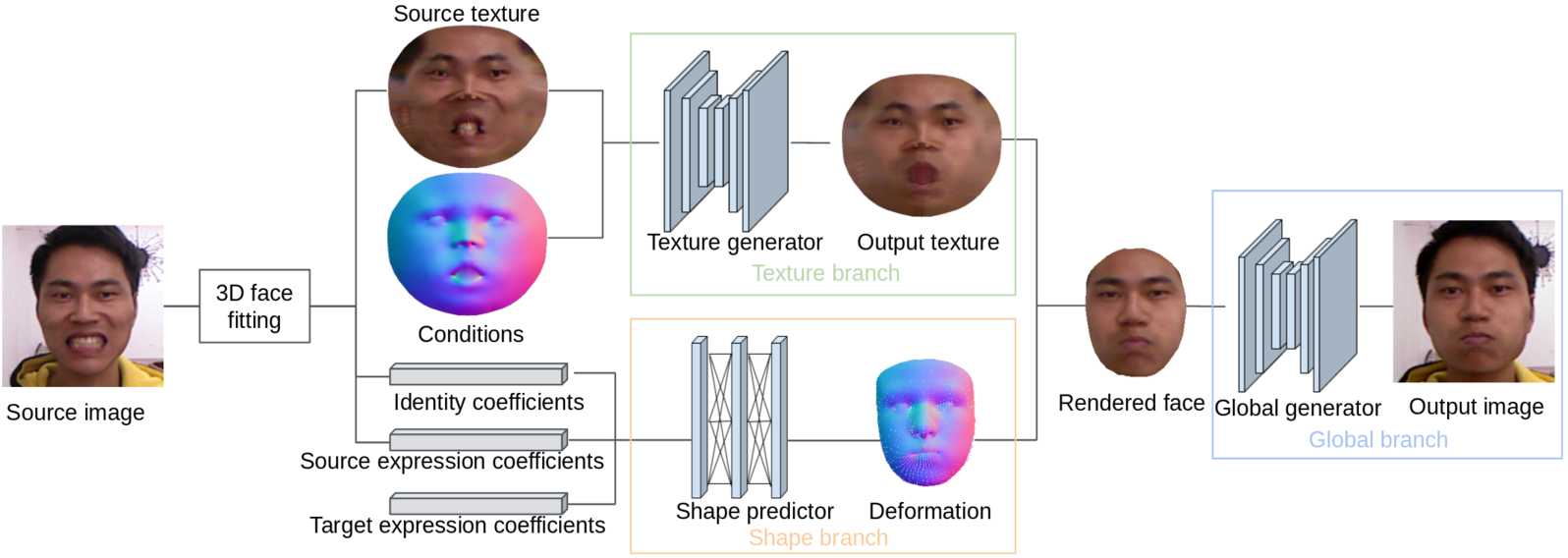}
    \caption{\textbf{Overview of our pipeline.} We first fit a 3DMM to the input and decouple it into the texture and shape coefficients. The texture branch assumes the source texture and the spatial representation of the target expression to produce the output shape. The shape branch uses the 3DDM coefficients to output a shape deformation. Finally, the global branch blends the two outputs in the image space. }
    \vspace*{-2mm}
    \label{fig:pipeline}
\end{figure*}

We compare the proposed method to the most recent face manipulation methods~\cite{StarGAN2018,GANImation2018} and show that our approach is superior in all the experiments.
In the user studies that we conducted, the presented method is preferred more than 85\% of the time when compared with the existing works. When compared to the ground truth testing images, our method is preferred in 53\% of cases, supporting that it is difficult for a human to distinguish real images from those generated by our method.

\section{Related Work}
\vspace*{-1mm}

We review relevant geometry based methods and deep generative methods for face manipulation.

\textbf{Geometry-based methods.} A pioneering work of Blanz and Vetter~\cite{3DMM1999} presented the first public 3D Morphable Model (3DMM). They densely captured surface geometry and color data of 200 identities and created a linear model to represent the face variations of different subjects using principal component analysis (PCA). Vlasic~\etal~\cite{Vlasic2005} proposed a multilinear model of facial expressions for tracking and re-targeting. Cao~\etal~\cite{cao2014facewarehouse} proposed FaceWarehouse, an extensive facial expression database, which contains 47 different facial expressions for each of the 150 subjects. This dataset later became one of the most adopted datasets for 3D face fitting and animation \cite{Cao:2014:DDE,Zhu2016}.

In~\cite{cao2014facewarehouse}, they first fitted a 3D face shape to match the input image, and then changed the expression coefficients to perform animation by warping the image to a new expression. 
Thies~\etal~\cite{face2face2016} presented Face2Face for real-time video-to-video facial expression re-targeting. They first fit a 3DMM together with lighting parameters and re-render it in the target video. 
Although these geometry-based methods produce convincing results of large-scale motions, they are unable to model parts not existing in the source image, such as teeth when the mouth is closed, and resort to rendering such parts using conventional graphics approaches. Therefore, these methods often fail to achieve realistic results, as humans are especially sensitive to non-realistic artifacts in faces.

\textbf{Deep generative methods.} Face manipulation can be viewed as the unpaired image-to-image translation problem~\cite{liu2018intriguing,CycleGAN2017} . Until very recently, one had to train a separate model, attribute-by-attribute to perform face manipulation~\cite{liu2017unsupervised}. Lample~\etal~\cite{lample2017fader} proposed to additionally control the intensity of the attribute. Their work can change two attributes at the same time, but only at the cost of reduced image quality. Choi~\etal~\cite{StarGAN2018} used conditional image-to-image translation to allow multiple attributes to be trained together in an unsupervised fashion. These attributes can include gender, age, hair color, expression and so on. Despite the impressive results, their approach is still limited to a finite number of attributes, preventing fine-grained manipulation. Several video generation methods for face animation were proposed. Given a face image, such methods perform video prediction~\cite{Tulyakov:2018:MoCoGAN} or motion transfer~\cite{siarohin2018animating,wiles2018x2face} to manipulate faces. Recently, Pumarola~\etal~\cite{GANImation2018} presented a work performing anatomically-aware face animation. Similarly to us, they animate faces according to Facial Action Units. 

The method presented in this paper is different than geometry-based and deep generative methods in that it combines the benefits of both lines of work in a single end-to-end trainable framework. As opposed to purely 3DMM-based methods and similarly to deep generative works, our framework features high quality face texture synthesis. In contrast to deep generative works, and similarly to 3DMMs-based methods, our approach can generate arbitrary number of facial expressions. A key difference with Pumarola~\etal\cite{GANImation2018} is that we learn to explicitly disentangle shape and appearance into different branches. This enables learning a rich face prior from our shape branch, and allows the texture branch to focus on synthesizing realistic images.

\section{Method}

Our pipeline is shown in \fig{pipeline}. The approach requires a face image and the desired expression encoded by coefficients. We first fit the 3D face shape and camera projection matrix from the image, with which we extract textures (Sec.~\ref{sec:face_fit}). Then, we input the texture and the target expression to the texture branch and generate the target texture containing the details of the desired expression (Sec.~\ref{sec:texture}). As the 3DDM-based shape representations are often inaccurate, we use a fully connected network in the shape branch to predict a more accurate shape for improved synthesis quality (Sec.~\ref{sec:shape}). The predicted texture and shape are then combined and rendered to obtain a target image. We then use the global branch network on the target image to further improve the quality (Sec.~\ref{sec:global}).

\begin{figure}[t]
    \centering
    \includegraphics[width=0.98\linewidth]{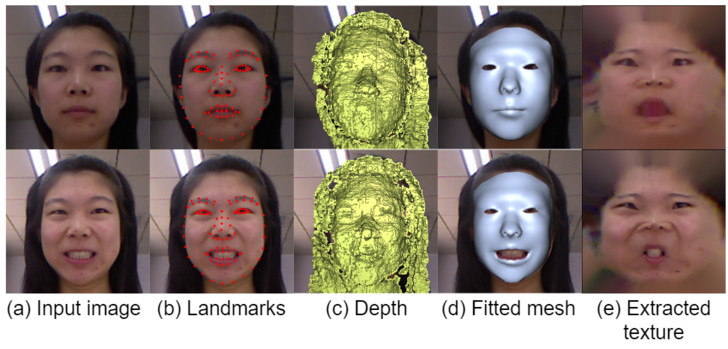}
    \vspace*{-5mm}
    \caption{Examples of fitting a 3DMM to an RGB-D image.}
    \label{fig:face_fit}
\end{figure}

\begin{figure*}[h]
	\centering
	\includegraphics[width=\linewidth]{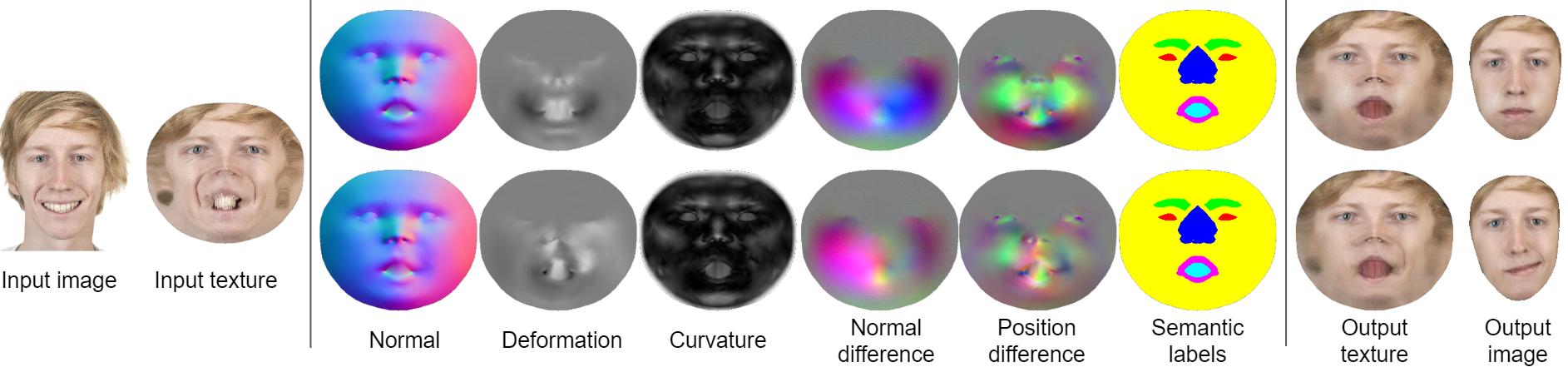}
	\caption{\textbf{Texture branch inputs and outputs.} The input image is mapped to the texture space. The texture branch uses important geometry information represented spatially using UV-maps. The output of the texture branch is a texture containing the desired target expression.}
	\label{fig:tex_branch}
    \vspace*{-5mm}
\end{figure*}

\subsection{3D Face Fitting}
\label{sec:face_fit}
Face fitting is the process of estimating the 3D face shape and the camera projection matrix given an input face image.
Following~\cite{cao2014facewarehouse}, we represent the 3D face shape using a bilinear model as:
\begin{equation}
    \mathbf{S} = C_r \times_2 \mathbf{a} \times_3 \mathbf{e},
\label{eqn:3DShape}
\end{equation}
where $\mathbf{S}\in \mathbb{R}^{3N}$ is the face shape, $N$ is the number of vertices, $C_\mathrm{r} \in \mathbb{R}^{3N\times N_\mathrm{a} \times N_\mathrm{e}} $ is the weight tensor, $\mathbf{a} \in \mathbb{R}^{N_\mathrm{a}}$ are the identity coefficients, $\mathbf{e} \in \mathbb{R}^{N_e}$ are the expression coefficients, $\times_i$ is the tensor contraction operation along the $i$th mode of the bilinear model. In our experiments, $N_\mathrm{e}=46, N_\mathrm{a}=50$ and $N=1220$.

Given a face image (\fig{face_fit}a), we first detect the $96$ 2D landmarks using \cite{KazemiS14} (\fig{face_fit}b). Then, we jointly estimate the camera projection transformation $\mathbf{M} : \mathbb{R}^{3}\rightarrow \mathbb{R}^2$, as well as identity and expression coefficients, by minimizing the L2 distance between the projected landmarks and detected landmarks. Note that we fix the identity coefficients for multiple images of the same person during optimization.

The inaccuracy in the fitting process causes the extracted textures to be misaligned and thus introduces additional variance for neural network to learn. To tackle this, we make use of the depth data when it is available. For the input image with a depth map (\fig{face_fit}c), we minimize the L2 distance between the shape vertices and its closest 3D depth points and then refine the shape using~\cite{Huang2006} (\fig{face_fit}d). When the depth is not available, we deform the shape to further reduce the landmark errors as in~\cite{cao2014facewarehouse}.

We define a 2D UV coordinate for each 3D shape vertex, consistent across the dataset. The textures are extracted with the UV coordinates, camera projection and fitted 3D shape using the standard rasterization pipeline.  (\fig{face_fit}e).

\subsection{Texture Branch} 
\label{sec:texture}

\newcommand{\T}{\mathbf{T}}
\renewcommand{\a}{\boldsymbol{\alpha}}

Our texture branch learns a function $G(\T^\mathrm{src},\mathbf{e}^\mathrm{src},\mathbf{e}^\mathrm{tgt})$ which transfers a texture $\mathbf{T}^\mathrm{src}$ extracted from the source image with the expression $\mathbf{e}^\mathrm{src}$, to texture $\mathbf{T}^\mathrm{tgt}$, containing the target expression $\mathbf{e}^\mathrm{tgt}$. Inspired by recent advances in image-to-image translation~\cite{pix2pix2016,CycleGAN2017}, we adopt conditional generative adversarial networks (cGAN) to learn the function $G$. 

\textbf{Input format.} 
Typically the generator $G$ is modeled as a convolutional neural network. In our case, the generator needs to take both the texture image $\mathbf{T}$ and the expression coefficients $\mathbf{e}$ as input. A straightforward approach to combine these different formats is to concatenate each element of $\mathbf{e}$ as a separate feature map to the input image $\mathbf{T}$ as in \cite{StarGAN2018,GANImation2018}. We argue that converting the geometry information of $\mathbf{e}$ into a spatial representation, such as a UV-map, helps better utilize local convolutional operations learned by the texture branch. 

In our implementation, this information includes object space normals, deformation, curvature, position difference, normal difference and semantic labels. We show examples in \fig{tex_branch}. Normal determines the local surface orientation which is considered important in shading. 
Deformation is determined by the ratio of the one-ring area near each vertex in the target and neutral expressions, where a small deformation value means compression and can be associated with wrinkles. 
Curvature differentiates bumped regions from flat regions. Position and normal differences imply similarities between source and target expressions near each vertex, indicating the likelihood of the output pixel resembling the input pixel at the same location. 
Furthermore, to address the translational equivariance issue of convolutions \cite{liu2018intriguing}, semantic labels are used to indicate different facial components which should be synthesized differently. These labels include eyes, eyebrows, nose, lips and inner mouth and others. As all the shapes have the fixed layout in the UV space, we manually define the labels on the 3D mesh and rasterize them to get the semantic map. We then use this semantic map for all the samples. We evaluate the effectiveness of our input format in Sec.~\ref{sec:ablate}.

\newcommand{\Treal}{\mathbf{T}_{i,p}^\mathrm{real}}
\newcommand{\Tfake}{\mathbf{T}_{i,p}^\mathrm{fake}}
\newcommand{\D}{\mathbf{D}}
\newcommand{\Lt}{L_2}
\newcommand{\Lb}{\bar{L_2}}
\newcommand{\x}{\mathbf{x}}
\renewcommand{\L}{\mathcal{L}}
\newcommand{\e}{\mathbf{e}}

\textbf{Loss functions.} 
Let $\Treal,\Tfake$ be the real and fake textures of identity $\mathbf{a}_i$ under the expression $\e_{p}$. We design three discriminator terms to improve the synthesis quality:
\begin{itemize} 
    \item $D_\mathrm{real}$ is the standard discriminator to distinguish between real textures $\Treal$ and synthesized fake textures $\Tfake$. 
    \item $D_\mathrm{pair}$ is used to ensure pair consistency between the texture and the expression coefficients as \cite{reed2016generative}. Our discriminator $D_\mathrm{pair}$ learns to differentiate matched pairs of real texture and expressions $(\Treal,\e_p)$ from matched pairs of fake texture and expressions $(\Tfake,\e_p)$ and mismatched pairs of real texture and expressions $(\Treal,\e_r)$,  where $\e_r$ is a random expression.
    \item  $D_\mathrm{iden}$ is designed to preserve identities. It is used to differentiate real textures with the same identity $(\mathbf{T}_{i,p}^\mathrm{real},\mathbf{T}_{i,q}^\mathrm{real})$, from real and fake textures with the same identity $(\mathbf{T}_{i,p}^\mathrm{real},\mathbf{T}_{i,q}^\mathrm{fake})$, and real textures with different identities $(\mathbf{T}_{i,p}^\mathrm{real},\mathbf{T}_{j,q}^\mathrm{real})$, where $p,q$ index random expressions and $i,j$ index different identities.
\end{itemize}

\begin{figure}[t]
    \centering
    \includegraphics[width=0.98\linewidth]{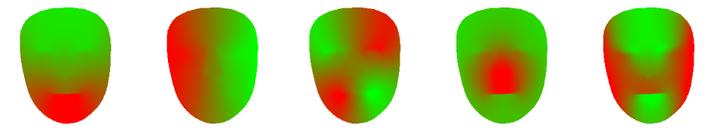}
    \caption{The first 5 eigenvectors of an average face model. In each eigenvector, the vertices with similar colors have similar deformation.}
    \label{fig:eigen_faces}
\end{figure}

We use LSGAN~\cite{MaoLXLW16} to calculate the respective loss terms $\L_\mathrm{real}, \L_\mathrm{pair}, \L_\mathrm{iden}$. The combined GAN objective writes as:
\begin{equation}
    \L_\mathrm{GAN}=\L_\mathrm{real}+\L_\mathrm{pair}+\L_\mathrm{iden}.
\end{equation}
The objective for our discriminators is
\begin{equation}
    \max_{D_\mathrm{real},D_\mathrm{pair},D_\mathrm{iden}} \L_\mathrm{GAN}.
\end{equation}

The generator $G$ minimizes $\L_\mathrm{GAN}$ and is supervised by $\L_1$ loss 
and perceptual loss~\cite{johnson2016perceptual} $\L_\mathrm{perc}$. Though the original perceptual loss is proposed in image space, we find it effective in texture space as well (Sec.~\ref{sec:ablate}).
Thus our generator objective is
\begin{align}
\min_{G} \L_\mathrm{GAN}+\lambda_{\L_1} \L_1 +\lambda_\mathrm{perc} \L_\mathrm{perc}.
\end{align}
In our experiments, we empirically set $\lambda_{\L_1}=10,\lambda_\mathrm{perc}=10$. For more details of our loss terms, please see our supplementary materials.

\subsection{Shape Branch}\label{sec:shape}
\renewcommand{\S}{\mathbf{S}}
\newcommand{\B}{\mathbf{B}}
\renewcommand{\D}{\mathbf{D}}

The 3D face shape $\S$ is a non-linear function of the expression coefficients due to the complex interaction of muscles, flesh and bones.
Previous works \cite{Cao:2014:DDE,cao2014facewarehouse} model this complex interaction linearly. 
Although this method is simple and widely adopted, we argue that these limited expression models can only represent the large-scale motion, and struggle to capture the fine-grained details.

To further increase the accuracy of the shape branch, we deform the face shape either through depth or landmarks as mentioned in Sec~\ref{sec:face_fit}. To fully capture these geometric details, we formulate the shape function as a linear part using Eqn.~\ref{eqn:3DShape} and and a non-linear part $D(\mathbf{a}, \mathbf{e}^\mathrm{src}, \mathbf{e}^\mathrm{tgt})$, which is an additional deformation field. Similarly to \cite{Tran2018}, we train a neural network to learn only the non-linear deformation $D$ to reduce variance.

The output of $D(\mathbf{a}, \mathbf{e}^\mathrm{src}, \mathbf{e}^\mathrm{tgt})$ represents the per vertext displacement vectors. These vectors can be very high dimensional. To reduce dimensionality,  we model the displacements with a spectral representation as in~\cite{Bouaziz2013}. More specifically, we compute eigenvectors of the $k$ smallest non-zero eigenvalues of the graph Laplacian matrix of a generic 3D face shape~\cite{Levy2009} and use them as the basis of vertex displacements. We use a fully connected network with 2 hidden layers to predict the basis coefficients. \fig{eigen_faces} shows the first 5 eigenvectors. In our experiments we set $k=100$.

\subsection{Global Branch} \label{sec:global}
\begin{figure}[t]
    \centering
    \includegraphics[width=0.98\linewidth]{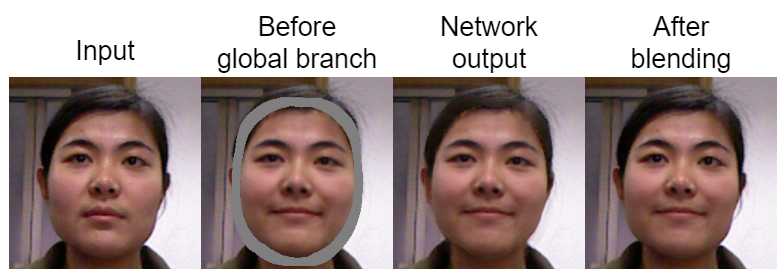}
    \vspace*{-3mm}
    \caption{Demonstration of the global branch.}
    \vspace*{-5mm}
    \label{fig:global_branch}
\end{figure}

We use the predicted texture $\hat{\T}$ and shape $\hat{\S}$ to render the predicted face on the image. The goal of the global branch is to blend this face into the background seamlessly. We show the process in \fig{global_branch}. We first make the artificial margin between the rendered face and the background and train a network to hallucinate in between. The margin is computed using a dilation approach with kernel size 12. To fill in the margin, one could use image inpainting techniques \cite{yu2018free,yu2018generative}. We have a simpler problem since the input image is usually similar to the background image. Therefore, we use the global network that takes the input image, the rendered face and the region outside of the margin as input. The network then learns to blend the generated face and the background together. Occasionally this still produces artifacts near the boundary. Therefore at test time, we apply image blending with the input image as a post-processing step. We describe this step in more details in the supplemental materials.

\section{Implementation Details}
\label{sec:impl}
\textbf{Datasets.} Our datasets include FaceWarehouse \cite{cao2014facewarehouse} and Chicago Face Dataset (CFD) \cite{ma2015chicago}. For the training set, we use 493 identities from FaceWarehouse, each with at least 20 different expressions and 152 identities from CFD, each with at least 5 expressions. Among this data, 140 identities in Facewarehouse have depth. For the test set, we use 87 identities from Facewarehouse and 5 identities from CFD. Our datasets span different genders and skin colors. We use $256\times256$ for our image and texture resolution. To further increase the resolution multistage generative models can be employed~\cite{karras2017progressive,zhang2017stackgan}.

\textbf{Network architecture.} The texture and global branch generators adopt pix2pix~\cite{pix2pix2016} architecture with attention maps~\cite{GANImation2018}. We change the transposed convolutions to upsampling layers followed by 3x3 convolutions. Similarly our discriminators adopt the pix2pix discriminator architectures. See the supplement for more details.

\textbf{Training.}
We use Adam \cite{Adam:KingmaB14} optimizer with a learning rate of 0.0001, $\beta_1 = 0.5$, $\beta_2 = 0.9$. We first train the texture branch and the shape branch. Then we fix their weights and train the global branch. We use a single NVIDIA Tesla V100 GPU and we train for 5 days to get the best results.

\section{Experiments}
In this section, we first conduct an ablation study to evaluate the design choices in our system. Next, we compare our approach with other approaches both qualitatively and quantitatively. Finally, we show additional qualitative results.

\subsection{Ablation Study}\label{sec:ablate}

\begin{figure}
    \centering
    \includegraphics[width=0.98\linewidth]{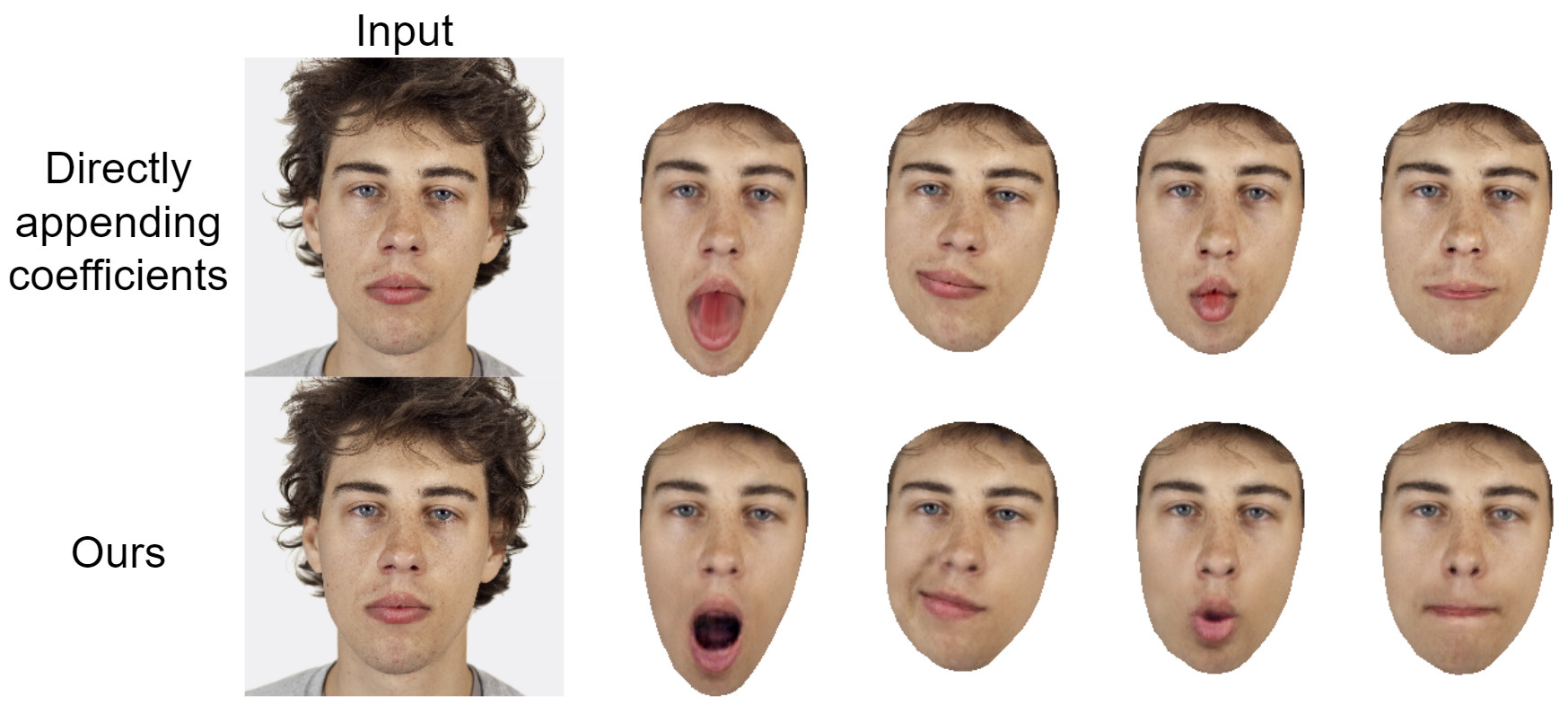}
    \vspace*{-2mm}
    \caption{Qualitative samples of different inputs of the texture branch evaluated on rare facial expressions. Note that the proposed approach generalizes better compared to the standard method of directly appending expression coefficients.}
    \vspace*{-2mm}
    \label{fig:ablate_cond}
\end{figure}

\textbf{Texture branch input format.} We compare our proposed input format with directly concatenating expression coefficients to the input of the neural network as in~\cite{StarGAN2018,GANImation2018}. We show that our approach generalizes better by transferring an image from CFD to a rare expression that CFD rarely covers in \fig{ablate_cond}. The model (top row) which appends expression coefficients directly as input fails to generate the correct appearance for regions like the inner mouth, cheeks near mouth corners and lips. This occurs since the generator has rarely seen the combination of this face skin color with these specific coefficients in the training dataset. The proposed approach, which conditions on the texture branch on the spatial representation of geometry information, generalizes better. We believe our approach better uses the local convoluational structure of the neural network.

\begin{figure}[t]
    \centering
    \includegraphics[width=1.0\linewidth]{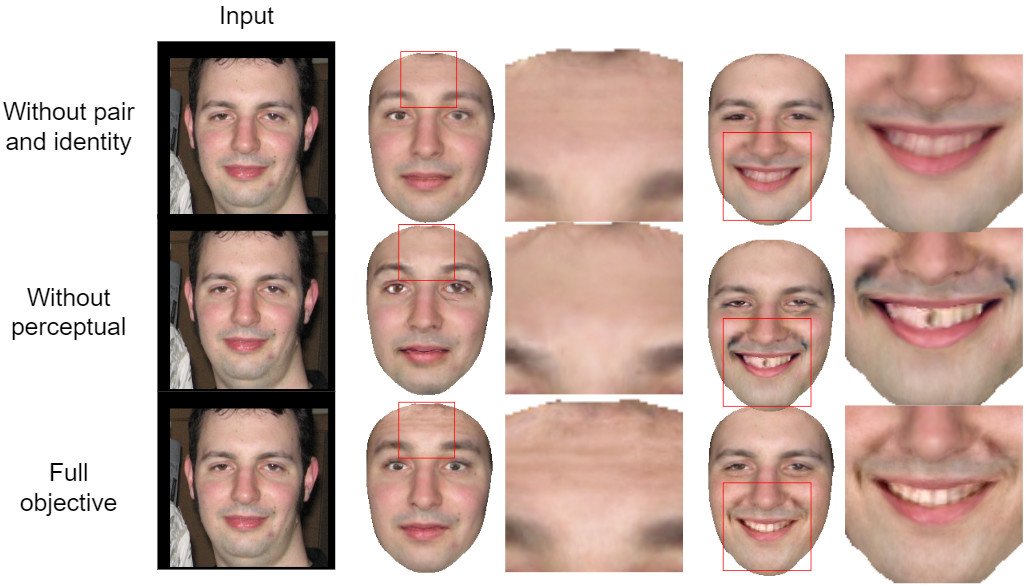}
    \vspace*{-2mm}
    \caption{Qualitative evaluation of different loss terms. The model trained with the full objects generates images with higher fidelity.}
    \vspace*{-2mm}
    \label{fig:ablate_loss}
\end{figure}

\textbf{Texture branch loss functions.}
We apply ablation study on our loss terms. We show results in \fig{ablate_loss}. We first remove $\L_\mathrm{pair}$ and $\L_\mathrm{iden}$. The $\L_\mathrm{pair}$ term is designed to enforce texture-expression pair consistency. We can observe that expression specific features such as wrinkles are less observable after the removal. $\L_\mathrm{iden}$ is designed to preserve identities, which helps direct appearance details from the source texture to the synthesized texture. We can see that the synthesized images, especially near the teeth region, are more blurry after the removal. We then remove the perceptual loss while keeping the rest unchanged. Similar to $\L_1$ loss~\cite{pix2pix2016}, the perceptual loss helps avoid artifacts near the mouth region. It also allows the network to capture more subtle details such as wrinkles on the forehead compared to $\L_1$ loss alone.

\begin{figure}
 \centering
    \includegraphics[width=1.0\linewidth]{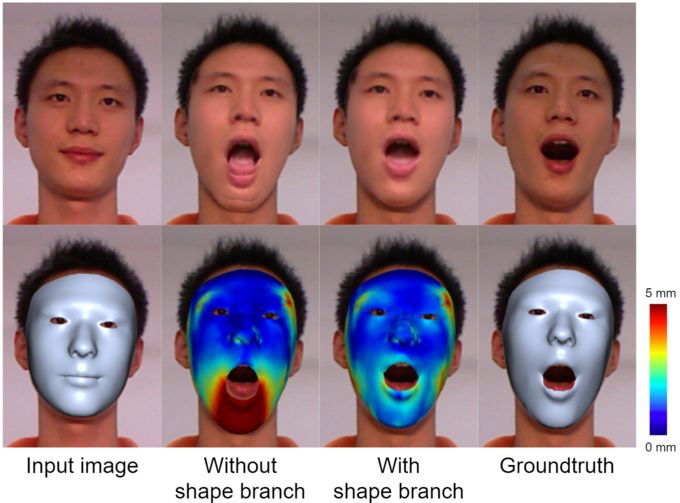}
    \vspace*{-2mm}
    \caption{Qualitative evaluation of the shape branch. The top row shows the input image, the generated images without and with the shape branch and the ground truth target image. The bottom row shows the fitted mesh with the color coded depth error in mm. Lower depth error makes the generated images more realistic.}
    \vspace*{-2mm}
    \label{fig:shape_compare}
\end{figure}

\textbf{Shape branch.}
We demonstrate that our shape branch generates more realistic shapes than the linear blendshapes both quantitatively and qualitatively. We first compute the root mean square error (RMSE) between the generated face mesh and our ground truth fitted face mesh in Table~\ref{table:shape_rmse}. After being deformed by our shape branch, the predicted mesh is closer to the ground truth.
\begin{table}
\centering
\caption{RMSE of vertices with/without shape branch. Lower number is better.}
\begin{tabular}{c|c}
\hline
 & RMSE (mm) \\
\hline
Without shape branch &  2.2158 \\
\textbf{With shape branch} &  \textbf{1.7619} \\
\hline
\end{tabular}
\label{table:shape_rmse}
\end{table}
We also show an example demonstrating the change of the mesh in \fig{shape_compare}. Without the shape branch, the fitted linear blendshapes tend to open the jaw more widely, which looks less natural, while our shape branch learns to close the jaw, such that the shape gets closer to the ground truth.

\begin{figure*}[t]
    \centering
    \includegraphics[width=1.0\linewidth]{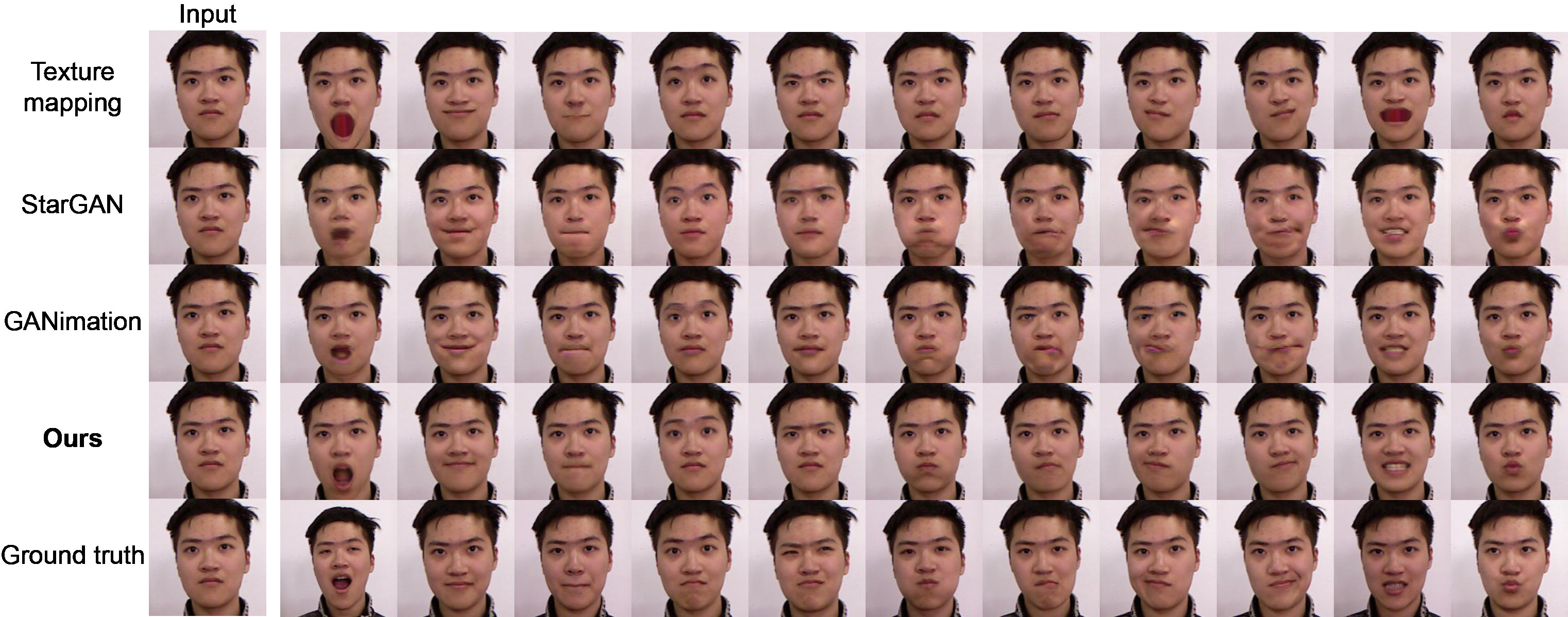}
    \vspace*{-2mm}
    \caption{\textbf{Comparison of face synthesis methods.} Given the same input image each method generated 11 different facial expressions. Our approach produced less artifacts and renders more realistically looking images than the competing methods.}
    \label{fig:method_compare}
    \vspace*{-2mm}
\end{figure*}

Note that despite the obvious benefits of the texture shape decoupling, our carefully designed input format, loss functions and shape branch are necessary for best results.

\subsection{Comparisons}
We compare our face manipulation results to the direct texture mapping approach~\cite{cao2014facewarehouse}, StarGAN~\cite{StarGAN2018} and GANimation~\cite{GANImation2018}. The method in \cite{cao2014facewarehouse} is a linear model combined with a computer graphics rendering approach, which also separates the texture and the shape but does not alter the texture. The latter two train on image space only concatenating the attributes or action units directly to the input. To evaluate the effectiveness of different methods at handling wide ranges of extreme expressions, we choose FaceWarehouse \cite{cao2014facewarehouse} as our training set because it contains many challenging expressions other datasets do not normally cover. We use the 87 identities in our test set as mentioned in Sec.~\ref{sec:impl} and we do not include CFD for easier comparisons. We trained StarGAN using $20$ different expressions as attributes. We implemented GANimation with all the attention mechanisms and loss terms, except that we replaced the regressor with a classifer in their discriminator, which tends to give better results on our dataset. For all the comparison experiments, we transfer neutral expressions to different expressions. We use the real captured data from \cite{cao2014facewarehouse} as the ground truth.

\textbf{Qualitative study.} 
We show several examples in \fig{method_compare}. Direct texture mapping is not able to generate wrinkles in the smiling expression, teeth in mouth opening expression or correct shading details in mouth blowing expression. For StarGAN and GANimation, we observe that they tend to produce more artifacts in expressions that have larger scale facial movements like mouth opening and mouth sloping. We hypothesize that this is because the competing approaches need to learn a complex model with all the rigid pose, shape and appearance variance together, while our fitting process and shape branch take the first two away, leaving a simpler function for the texture branch to learn. We also find that GANimation sometimes leaves the details from the input image in the output. Interested readers can magnify the lips region on the 4th column and eyebrows region in the 5th column to see the artifacts. We hypothesize that this is a problem caused by the attention mechanism in the image space. Our approach has a fixed texture layout and thus does not have this problem.  

Note that our synthesized images have different camera poses than the ground truth. This is because FaceWarehouse is captured with different head poses and our images are cropped differently based on the face sizes. Also note that the synthesized images look different from the ground truth. This is because there are numerous ways that a person can perform an expression and our method only generates a possible realization of that expression.

\begin{table}[t]
\centering
\caption{Quantitative comparisons and user studies results of different methods. We report the Average Content Distance (ACD, lower is better) and the user preference score (higher is beter). Best results in bold.}
\begin{tabular}{c|c|c}
\hline
 \multirow{2}*{Methods}
            & \multirow{2}*{ACD}
                        & User Preference    \\
            &           & Ours / Others        \\    
\hline
Texture mapping \cite{cao2014facewarehouse} &  0.6194 & \textbf{69.8} / 30.2\\
StarGAN \cite{StarGAN2018} & 0.5981 & \textbf{86.8} / 13.2\\
GANimation \cite{GANImation2018} & 0.5595 & \textbf{86.2} / 13.8\\
\textbf{Ours}  & \textbf{0.5107} & N/A\\
Ground truth & 0.4608 & \textbf{53.4} / 46.6 \\
\hline
\end{tabular}
\vspace*{-4mm}
\label{table:comparisons}
\end{table}

\textbf{Quantitative study.}
We adopt Average Content Distance (ACD) from \cite{Tulyakov:2018:MoCoGAN} to evaluate how well identities are preserved using different methods. We extract feature vectors from each synthesized image and compute the $\L_2$ distance to the feature vector of the input image. We show the results in Table~\ref{table:comparisons}. Our method gives the best results besides ground truth. Note that we do not optimize with respect to any pretrained face recognition networks at training time. We attribute our lower ACD to our disentanglement representation of texture and shape, which makes it easier to preserve identities.

\textbf{User study.} We perform a user study on Amazon Mechanical Turk (AMT), where each worker is presented with the reference image, an image synthesized by our method and an image synthesized by a competing method. We ask the turkers to evaluate the synthesized images based on their quality, realism of the expression, and similarity to the reference image. Since faces in ground truth images have different poses, for comparison with ground truth we only ask the subjects to evaluate based on the quality of the image and expression, eliminating other irrelevant factors as much as possible. For each comparison, we have $1,740$ pairs of images and each pair is evaluated by 3 workers. We only accepted turkers with a lifetime HIT approval rate $\geq 95\%$. We show the results in Table~\ref{table:comparisons}. Users prefer our methods over all other methods. We get a slightly higher preference score than ground truth. This proves that it is difficult for humans to distinguish between the images generated by our method and the ground truth.

\subsection{More Results}
\begin{figure}
    \centering
    \includegraphics[width=1.0\linewidth]{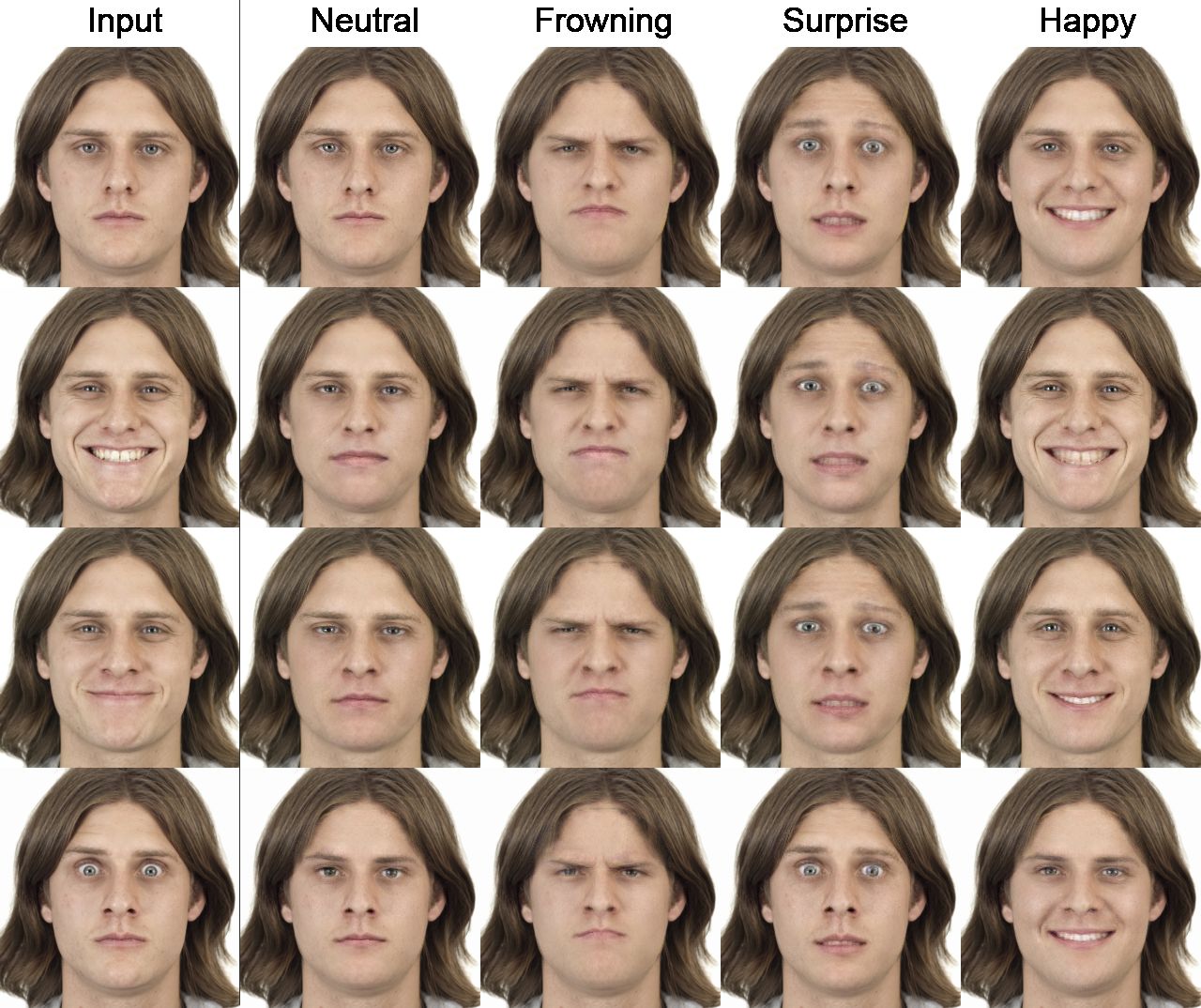}
    \caption{Manipulating images with different source expressions. Although the input images are different, each manipulated image looks plausible. }
    \label{fig:multi_expr}
    \vspace*{-5mm}
\end{figure}

\textbf{Different input expressions.} 
Our method can handle input expressions that are not neutral. We show synthesized images using the same person with different expressions as input in \fig{multi_expr}. Although our input images are different, our synthesized images with the same target expressions still look similar. We also note that the method can generate a different version of each expression for each subject.

\textbf{Images in the wild.}
We show examples of our method applied to images in-the-wild in \fig{teaser} and refer the reader to the supplementary materials for more in-the-wild results.
Due to the decoupled face representation and separate texture and shape branches, our method is robust to different identities, expressions, head poses or lighting.

\section{Conclusion}
We presented a 3D guided fine grained face manipulation approach to transfer from one arbitrary expression to another arbitrary expression. The method decomposes an image into shape and texture spaces, followed by processing of these spaces with separate branches. We showed the benefits of such a scheme. Conditioning the pipeline on the spatial representation of important geometry information is advantageous over the straightforward approach of directly appending expression coefficients. To further boost the quality, we introduced several of the loss functions accounting for the pairwise consistency and identity. Our ablation studies supported the proposed framework. Furthermore, our method showed a significantly better ACD score as well as a preference by human annotators when compared to the competing approaches. Finally, when compared to the real images, the annotators were not able to distinguish our generated images from the real images, fully supporting the benefits of the presented method.

\vspace*{3mm}

\textbf{Acknowledgements.} This work was mainly done when the first author Zhenglin Geng, who interned at Snap Inc. We also thank Davis Rempe, Rahul Sheth and Aletta Hiemstra for their help with the paper revision.

{\footnotesize
\bibliographystyle{ieee}
\bibliography{ref}
}

\newpage





\appendix
\renewcommand{\thesection}{\Alph{section}}
\section*{Appendices}
\begin{figure*}
\begin{center}
    \centering
    \includegraphics[width=1.0\linewidth]{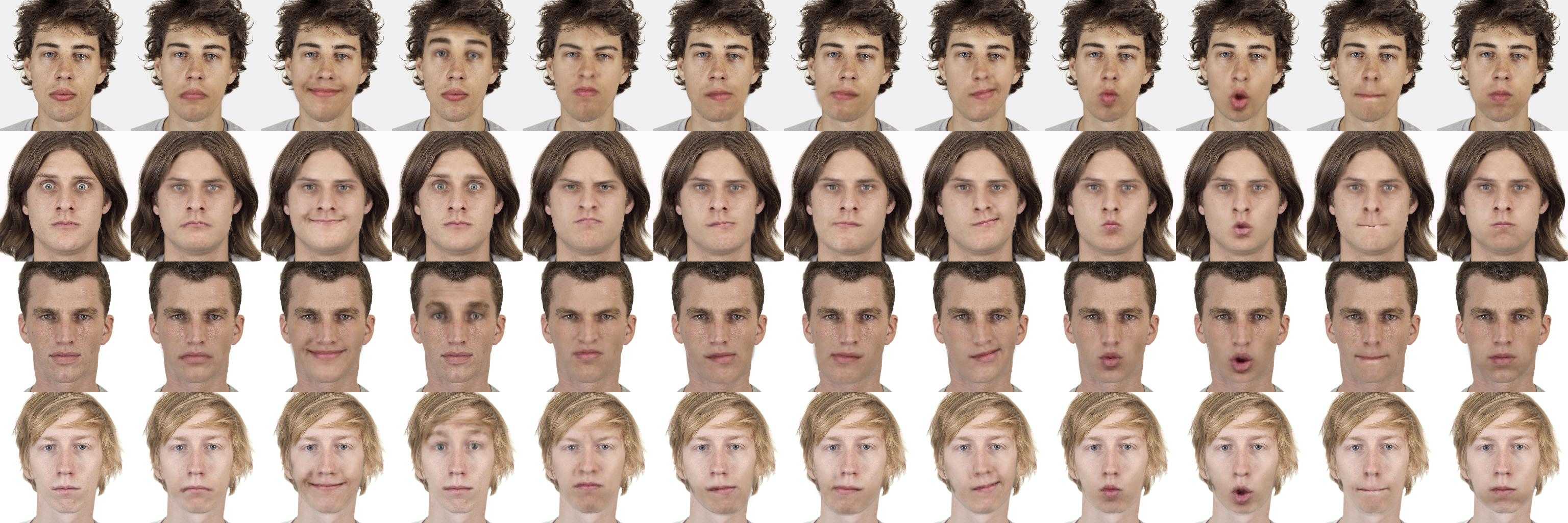}
    \captionof{figure}{More results from CFD. The first column are the input images. }
    \label{fig:cfd}
\end{center}
\end{figure*}
\section{More Results From CFD}
We present more results from CFD test set in Figure~\ref{fig:cfd}.

\section{More Results From FaceWarehouse}
\begin{figure*}
\begin{center}
    \centering
    \includegraphics[width=1.0\linewidth]{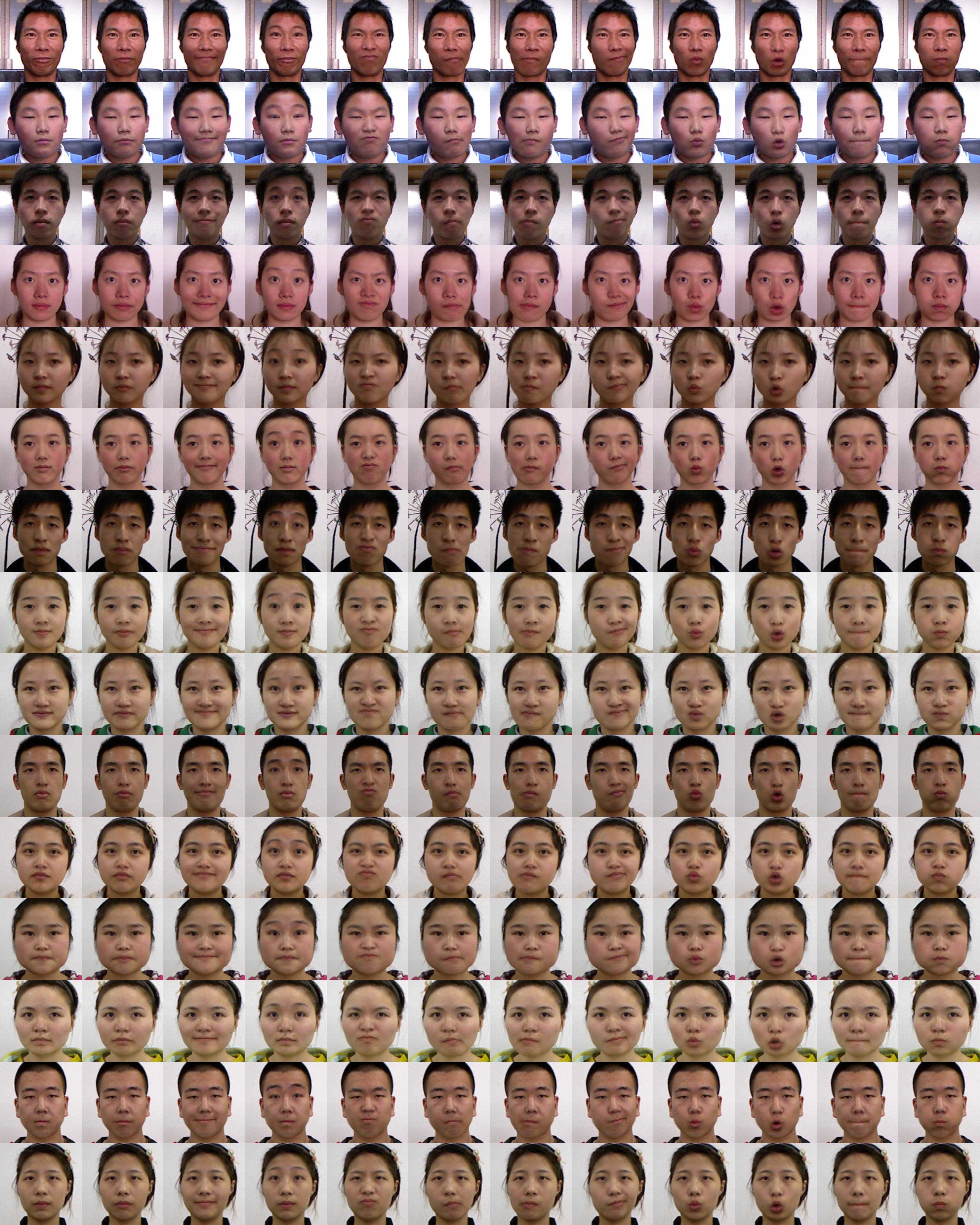}
    \captionof{figure}{More results from FaceWarehouse. The first column are the input images. }
    \label{fig:fw}
\end{center}
\end{figure*}

We present more results from Facewarehouse test set in Figure~\ref{fig:fw}.
 
\section{More Results from Images in the Wild}\label{sup:itw}
\begin{figure*}
\begin{center}
    \centering
    \includegraphics[width=1.0\linewidth]{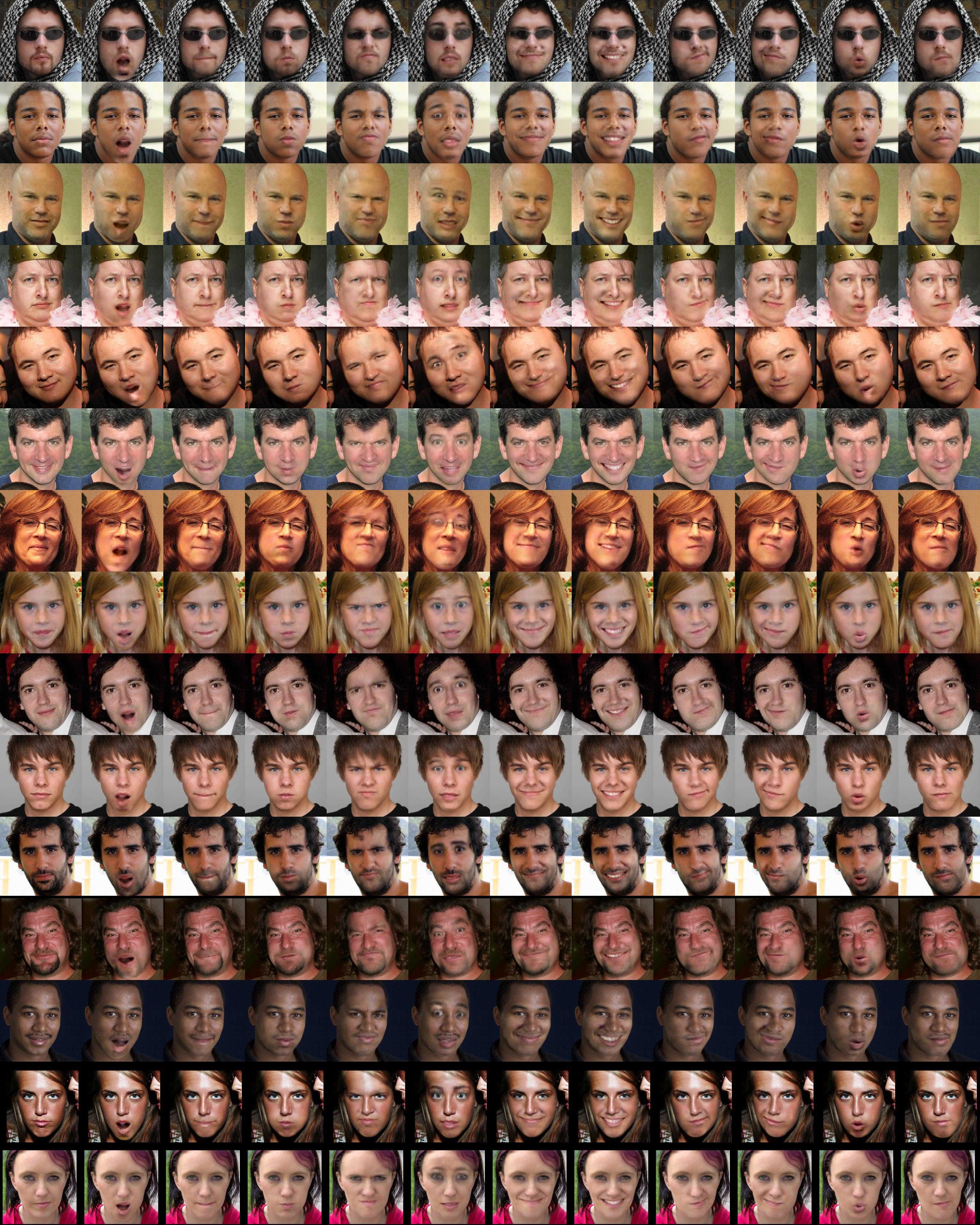}
    \captionof{figure}{More results from in-the-wild images. Note that our training set only contains images captured with frontal faces and good lighting, our trained model can work on some challenging in-the-wild images as well. }
    \label{fig:itw}
\end{center}
\end{figure*}

Though our training set only contains images captured in a lab setting with frontal faces and uniform lighting, we show that our trained model can work on more challenging in-the-wild images in Figure~\ref{fig:itw}.

\section{More Results on Continuous Editing}
As we use expression coefficients as conditions, we can trivially manipulate faces continuously. We show more results on continuous editing in the submitted video.

\renewcommand{\Lb}{\bar{L}_2}
\newcommand{\Lreal}{\L_{\mathrm{real}}}
\newcommand{\Lpair}{\L_{\mathrm{pair}}}
\newcommand{\Liden}{\L_{\mathrm{iden}}}
\newcommand{\Dreal}{D_{\mathrm{real}}}
\newcommand{\Dpair}{D_{\mathrm{pair}}}
\newcommand{\Diden}{D_{\mathrm{iden}}}
\section{More Details in $\Lreal,\Lpair$ and $\Liden$}\label{sup:loss}
As mentioned in the paper, our min-max game objective is composed of three terms: the realism term $\Lreal$, the pair-wise term $\Lpair$ and the identity term $\Liden$. These three loss terms are calculated from three discriminators $\Dreal$, $\Dpair$ and $\Diden$ respectively using LSGAN\cite{MaoLXLW16}. Let $\Treal,\Tfake$ be the real and fake textures of identity $\mathbf{a}_i$ under the expression $\e_{p}$. Let $\Lb(\x)=\parallel \x-1 \parallel ^2,\Lt(\x)=\parallel \x \parallel^2$, our loss terms are calculated as follows:
\begin{itemize} 
    \item $\Lreal$ is used to differentiate real images and fake generated images:
    \begin{equation}
    \Lreal=\Lb(\Dreal(\Treal))+\Lt(\Dreal(\Tfake)).
    \end{equation}
    \item $\Lpair$ is used to differentiate matched pairs of real texture and expressions $(\Treal,\e_p)$ from matched pairs of fake texture and expressions $(\Tfake,\e_p)$ and mismatched pairs of real texture and expressions $(\Treal,\e_r)$,  where $\e_r$ is a random expression:
    \begin{align}
    \Lpair&=2\Lb(\Dpair(\Treal,\e_p))  \\
    &+\Lt(\Dpair(\Tfake,\e_p)+\Lt(\Treal,\e_r)). \nonumber
    \end{align}
    where we multiply the first term by 2 to prevent the discriminator from simply producing a small value.
    \item  $\Liden$ is used to differentiate real textures with the same identity $(\mathbf{T}_{i,p}^{\mathrm{real}},\mathbf{T}_{i,q}^{\mathrm{real}})$, from real and fake textures with the same identity $(\mathbf{T}_{i,p}^{\mathrm{real}},\mathbf{T}_{i,q}^{\mathrm{fake}})$, and real textures with different identities $(\mathbf{T}_{i,p}^{\mathrm{real}},\mathbf{T}_{j,q}^{\mathrm{real}})$, where $p,q$ index random expressions and $i,j$ index different identities.
    \begin{align}
    \Liden&=2\Lb(\Diden(\mathbf{T}_{i,p}^{\mathrm{real}},\mathbf{T}_{i,q}^{\mathrm{real}}))\\
     &+\Lt(\Diden(\mathbf{T}_{i,p}^{\mathrm{real}},\mathbf{T}_{i,q}^{\mathrm{fake}}))+\Lt(\mathbf{T}_{i,p}^{\mathrm{real}},\mathbf{T}_{j,q}^{\mathrm{real}}). \nonumber
    \end{align}
\end{itemize}
\section{Network Architectures}\label{sup:net}
For our texture branch generator, we use pix2pix \cite{pix2pix2016} and attention map \cite{GANImation2018}. For our texture branch input, we concatenate input texture(3), normal(3), area deformation(1), curvature(1), normal difference(3), position different(3) and noise (1) together, thus the total number of channels for our input is 15. For our output, we use a separate attention map for each R,G,B channel, therefore the number of channels for our output is 6. To avoid being saturated in the gradients, we do not use any sigmoid or tanh activations. For the hidden layers in the middle, we use the UNet structure with skip link. For the encoder, we use convolutional layers with filter size 4, stride 2 and padding 1 for downsampling. For the decoder, we use bilinear upsampling followed by a convolutional layer with filter size 3, stride 1 and padding 1 for upsampling. Following the notation from \cite{pix2pix2016}, we use $Ck$ denote Convolution-BatchNorm-ReLU layer with $k$ filters. \newline
\textbf{encoder:} \newline
$C32-C64-C64$ \newline
\textbf{decoder:} \newline
$C64-C32-C6$ \newline
All ReLUs in the encoder are leaky, with slope 0.2. All ReLUs in the decoder are not leaky.

\section{More Details in the Global Branch}
The goal of the global branch is to blend the rendered image seamlessly into the background. We first generate an margin by calling the OpenCV dilate function with a kernel size of 12. The our global branch takes the rendered face, input image and regions outside of the margin as input to hallucinate inside. Sometimes there is still an observable boundary in which case we apply image blending. We blend the image based on the vertex distance $d$ from the source expression mesh to the target expression mesh. The blending alpha is determined heuristically as $\exp(d^2/4)$.

\section{More Comparison with Texture Mapping }
\begin{figure}[h]
    \centering
    \includegraphics[width=0.98\linewidth]{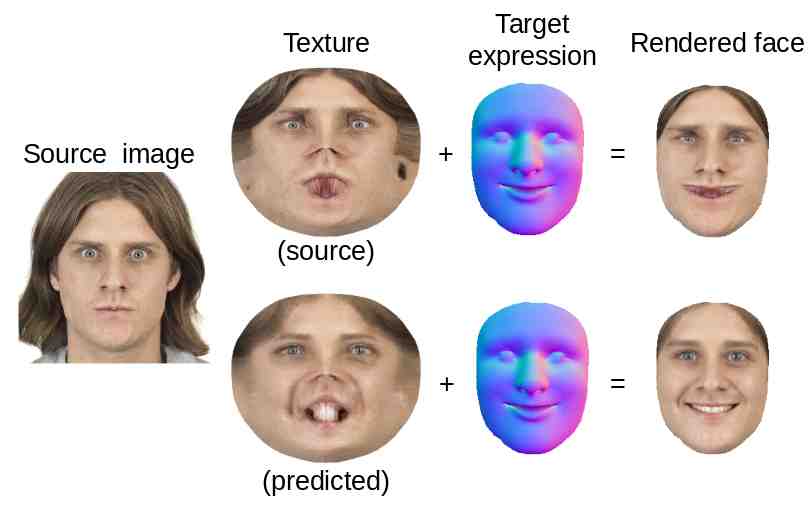}
    \caption{Demonstration of texture branch. This example shows transferring to a smiling expression. \textbf{Top}: the direct texture mapping approach \cite{cao2014facewarehouse}. \textbf{Bottom}: rendering using our predicted texture with the same smiling shape.}
    \label{fig:ablatex_tex}
\end{figure}

We show more examples of the difference between our approach and direct texture mapping approach. To manipulate expression in the image, one can change only the underlying shape without substantially changing the texture like \cite{cao2014facewarehouse}. However this can result in many artifacts, especially when the source and target expressions significantly differ. For example, in Fig.~\ref{fig:ablatex_tex}, if one directly uses the texture extracted from the source image and renders it with a smiling face shape, the missing crease and teeth and image distortion make the result less realistic. Our texture branch learns to reconstruct these missing parts and change the local appearance near the eyes, which makes the resulting image look natural. 

\newpage



\end{document}